\documentclass{article}

\usepackage{microtype}
\usepackage{graphicx}
\usepackage{subfigure}
\usepackage{booktabs} 

\usepackage{hyperref}

\usepackage[accepted]{synsml2023}
\usepackage[skip=0pt]{caption}
\usepackage{amsmath}
\usepackage{amssymb}
\usepackage{mathtools}
\usepackage{amsthm}

\usepackage[capitalize,noabbrev]{cleveref}

\theoremstyle{plain}

\theoremstyle{definition}

\theoremstyle{remark}

\usepackage[disable,textsize=tiny]{todonotes}

\synsmltitlerunning{
RANS-PINN based Simulation Surrogates for Predicting Turbulent Flows
}

\begin{document}

\twocolumn[
\synsmltitle{
RANS-PINN based Simulation Surrogates for Predicting Turbulent Flows
}
%
%
%
%



\begin{synsmlauthorlist}
\synsmlauthor{Shinjan Ghosh}{comp}
\synsmlauthor{Amit Chakraborty}{comp}
\synsmlauthor{Georgia Olympia Brikis}{comp}
\synsmlauthor{Biswadip Dey}{comp}
\end{synsmlauthorlist}

\synsmlaffiliation{comp}{Siemens Technology, Princeton, NJ 08536, USA}

\synsmlcorrespondingauthor{Shinjan Ghosh}{shinjan.ghosh@siemens.com}
\synsmlcorrespondingauthor{Biswadip Dey}{biswadip.dey@siemens.com}

\synsmlkeywords{Machine Learning}

\vskip 0.3in
]



\printAffiliationsAndNotice{}  

\begin{abstract}
Physics-informed neural networks (PINNs) provide a framework to build surrogate models for dynamical systems governed by differential equations. During the learning process, PINNs incorporate a physics-based regularization term within the loss function to enhance generalization performance. Since simulating dynamics controlled by partial differential equations (PDEs) can be computationally expensive, PINNs have gained popularity in learning parametric surrogates for fluid flow problems governed by Navier-Stokes equations. In this work, we introduce RANS-PINN, a modified PINN framework, to predict flow fields (i.e., velocity and pressure) in high Reynolds number turbulent flow regimes. To account for the additional complexity introduced by turbulence, RANS-PINN employs a 2-equation eddy viscosity model based on a Reynolds-averaged Navier-Stokes (RANS) formulation. Furthermore, we adopt a novel training approach that ensures effective initialization and balance among the various components of the loss function. The effectiveness of the RANS-PINN framework is then demonstrated using a parametric PINN.
\end{abstract}
\section{Introduction}
The traditional approach to designing complex devices and systems, for example, aerodynamic surfaces and thermal management systems, involves a back-and-forth interplay between exploring the design and operating space and assessing performance through computationally intensive computational fluid dynamics (CFD) simulations. However, the high computational cost associated with high-fidelity CFD solvers like Simcenter Star-CCM+ or Ansys Fluent undeniably curtails the overall scope of the design optimization process, often leading to suboptimal design choices. In this context, neural networks, with their expressiveness to capture pertinent functional relationships between initial/boundary conditions and the solution field of a PDE and the ability to predict simulation outcomes by invoking a single forward pass, offer an excellent tool for building fast and accurate surrogate models for CFD simulations. Such deep learning based approaches can accelerate design evaluations significantly, facilitating the generation of enhanced design choices through fast predictions of simulation outcomes.

In recent years, there has been considerable attention given to the use of deep learning methods to expedite CFD simulations and thereby improve engineering design processes \citep{vinuesa2022enhancing, warey2020data, zhang2022demystifying}. While some approaches use deep learning to accelerate traditional CFD solvers \citep{hsieh2019learning, doi:10.1073/pnas.2101784118}, a certain body of research treats the flow problems as problems defined over a cartesian grid or an irregular mesh and uses techniques involving convolutional or graph neural operators to predict the flow fields \citep{hennigh2017lat, jiang2020meshfreeflownet, wang2020towards}. Alternatively, in another line of work, physics-informed neural networks (PINNs) exploit \textit{automatic differentiation} and incorporate the underlying PDEs to approximate the solution field \citep{raissi2019physics, white2019fast, nabian2020physics, zhang2020frequencycompensated, jin2021nsfnets}. In addition, self-supervised learning methods for solving PDEs with PINNs have also been explored \cite{dwivedi2019distributed, lu2019deepxde, nabian2019deep}. This expanding body of research demonstrates the ability of ML-based approaches to accurately predict simulation outcomes, such as flow and temperature profiles over a spatiotemporal domain, utilizing both mesh-based and mesh-free techniques. Notably, the inclusion of physics-based regularization in these formulations have proven instrumental in enhancing the quality of the results. 

PINNs combine differential equations, such as compressible and incompressible Navier-Stokes equations, with experimental data or high-fidelity numerical simulations. While their ability to replace existing CFD solvers is a matter of debate, PINNs can accelerate simulations \citep{kochkov2021machine}, reconstruct flow domains from a limited sensor or experimental data \citep{wang2022dense}, and create parametric surrogates for design exploration and optimization \citep{GAPINN2022, SUN2023116042}. However, current PINN methods encounter challenges due to the complex interaction among the individual components of the loss function (both supervised and unsupervised), particularly when dealing with high-dimensional, non-convex PDE based losses. These challenges become more pronounced as the physics of the problem becomes more intricate, e.g., turbulent flows. RANS, the most commonly used turbulent CFD simulation tool, offers reasonably accurate solutions at a lower computational cost compared to high-fidelity \textit{direct numerical simulation} (DNS) and \textit{large eddy simulation} (LES), which require even finer mesh refinement to adequately capture all turbulence scales, further increasing computation time. Since its introduction by \citet{LAUNDER1974269}, $k$-$\epsilon$ model ($k$ is the turbulent kinetic energy and $\epsilon$ is the turbulent dissipation rate) has been established as a preferred model for efficient computation and real-world problems \citep{yang1993new, scott2004k, ghosh}. 

From this perspective, incorporating RANS-based turbulence modeling can significantly expand the application of PINNs in real-world simulation and design problems. However, using PINNs for RANS-based turbulence modeling is yet to be thoroughly studied \citep{en16041755}. Previous research by \citet{eivazi2022physics} employed RANS within PINNs but utilized a Reynolds-stress formulation instead of a 2-equation model like $k$-$\epsilon$. In contrast, \citet{xu2021explore} employed a PINN with RANS formulation to calculate missing flow components. In this study, we focus on constructing PINN-based surrogate models for turbulent flow problems using a RANS formulation, specifically the $k$-$\epsilon$ model, along with relevant data. We refer to the resulting solution as RANS-PINN and implement it using Nvidia Modulus (22.03) \citep{ModulusDocu, hennigh2021nvidia}. The proposed training regime first pre-trains the network using data losses and then introduces the physics losses in a carefully crafted manner. We first assess RANS-PINN on three distinct geometries: a cylinder, an airfoil, and flow over backward facing step; and, then employ it to learn a parametric PINN for flow over a cylinder. This approach improves upon the existing turbulence modeling capabilities of Nvidia Modulus, while also adding to the very few existing studies on RANS based turbulence modeling using PINNs. 
\begin{figure*}[t]
\centering
\includegraphics[width=0.85\textwidth]{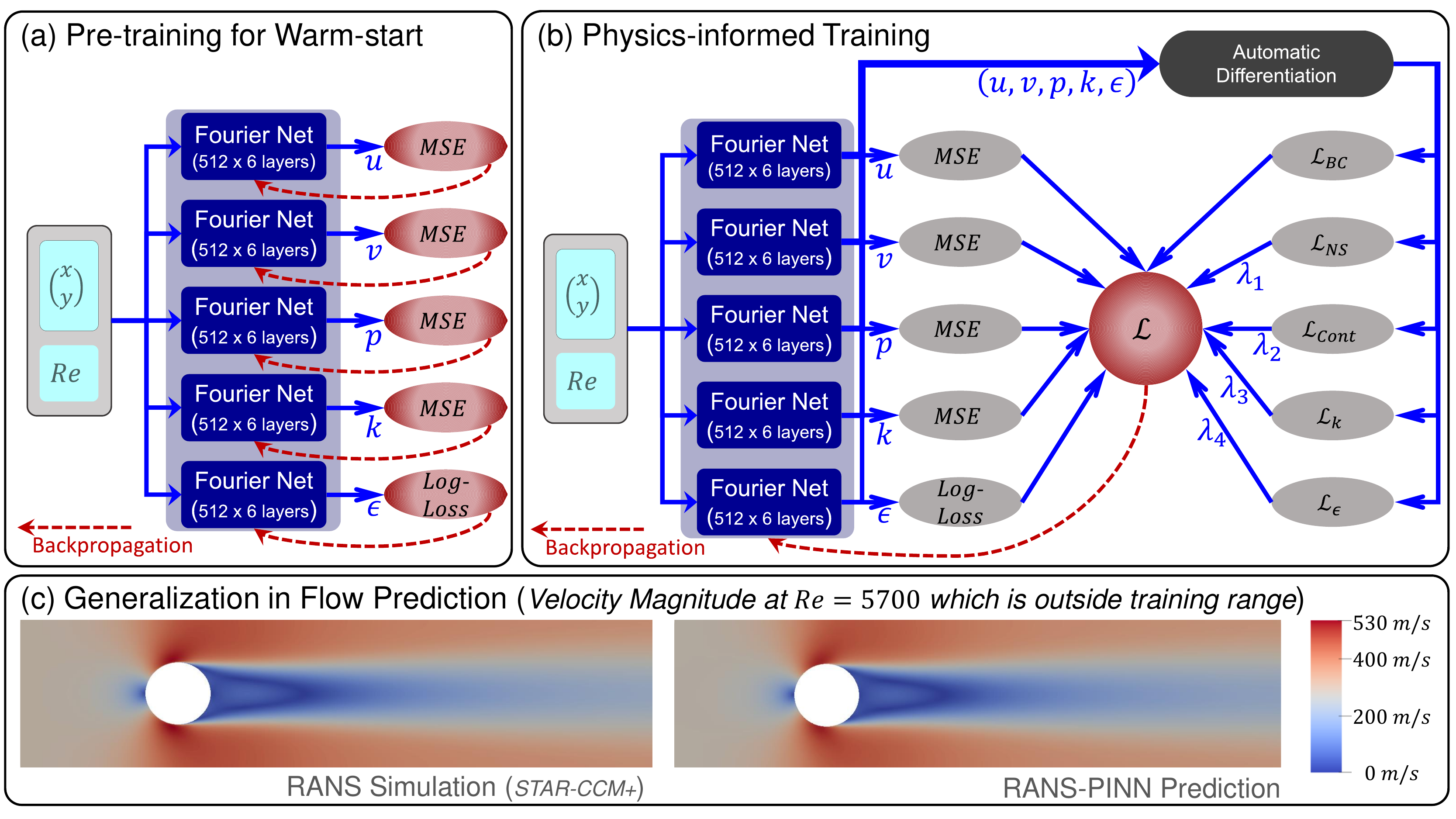}
\caption{\small{RANS-PINN framework for learning surrogates to predict turbulent flow.}}
\label{arch_PINN}
\end{figure*}
\section{RANS-PINN}
\subsection{Governing physics}
The underlying physics is governed by the continuity equation (to \textit{conserve mass}), Navier-Stokes equation (to \textit{conserve momentum}), and standard $k$-$\epsilon$ turbulence model. By letting $u$ and $p$ denote the flow velocity and pressure, respectively, continuity and Navier-Stokes equation can be expressed as:
\begin{align*}
\textrm{\textbf{NS:}}\quad & 
\nabla(u) = 0
\\
\textrm{\textbf{Cont:}}\quad &
\rho(u\cdot\nabla)u + \nabla(p)-  \mu_{eff}\nabla^{2} u = 0,
\end{align*}
where $\rho$ is density of the fluid, $\nabla$ denotes the vector differential operator, and $\mu_{eff}:= \mu + \mu_t= \mu+0.09k^2/\epsilon$ represents the effective viscosity, i.e., the sum of molecular viscosity ($\mu$) and turbulent viscosity ($\mu_t$). In addition, the $k$-$\epsilon$ turbulence model can be expressed as:
\begin{align*}
\textrm{\textbf{$k$:}}\quad &
\nabla (\rho u k)
=
\nabla\left[\left( \mu+\frac{\mu_{t}}{\sigma_{k}} \right)\nabla k\right] + P_{k} - \epsilon
\\ 
\textrm{\textbf{$\epsilon$:}}\quad &
\nabla (\rho u \epsilon)
= 
\nabla\left[\left(\mu+\frac{\mu_{t}}{\sigma_{\epsilon}}\right)\nabla\epsilon\right] + (C_{1}P_{\epsilon}+C_{2}\epsilon)\frac{\epsilon}{k}
\end{align*}
where, $C_1 = 1.44$, $C_2 = 1.92$, $\sigma_k = 1$, and $\sigma_{\epsilon} = 1.3$ are empirical model constants. In addition, $P_k$ and $P_\epsilon$ are production terms. The \textit{Reynolds number} for this system is defined as: $Re= \rho u_{inlet} L/\mu$, where $u_{inlet}$ is the inlet velocity and $L$ is the characteristic length.
\subsection{RANS-PINN architecture and training regime}
The RANS-PINN architecture (Fig.~\ref{arch_PINN}) uses Fourier neural operators \citep{li2021fourier} with their default hyperparameters used in Modulus \citep{ModulusDocu}. For each of the individual output variables (i.e., $u$, $p$, $k$, and $\epsilon$), we use separate neural networks, all sharing the same input variables consisting of positional coordinates $(x,y)$ and the associated Reynolds number. These networks are connected to the supervised/data loss, as well as the nodes of the PDE loss components.

Conventional approaches to training PINNs involve introducing data and PDE losses simultaneously at the start of the training phase, often with equal weight multipliers. However, they often results in noisy training losses, slow convergence, and high validation error. RANS-PINN addresses these challenges by employing a pre-training step that only uses the data-driven supervised loss. During pre-training, each of the individual networks is updated independently using their corresponding data loss. Following pre-training, we introduce the PDE constraints into the loss function. Moreover, to normalize the effect of the individual components of the PDE loss function, we scale them by the inverse of their corresponding residual values. We then use \textit{Adam} with a decaying step size (with the initial step size of 0.001 and a decay rate of 0.95) until the training loss converges. 

To address the challenges associated with abrupt changes observed in the turbulence dissipation term $\epsilon$ near wall and free shear regions, we use a \textit{logarithmic loss function} for both data and PDE losses associated with $\epsilon$. Everything else is computed using an \textit{MSE loss function}. The overall loss function can then be expressed as:
\begin{equation}
\mathcal{L} = \mathcal{L}_{data} + \mathcal{L}_{BC} + \mathcal{L}_{PDE},
\end{equation}
where the PDE loss is defined with weights $\lambda_i$'s as:
\begin{equation}
\mathcal{L}_{PDE} = 
\lambda_1\mathcal{L}_{NS} + \lambda_2\mathcal{L}_{Cont} + \lambda_3\mathcal{L}_{k} + \lambda_4\mathcal{L}_{\epsilon}.
\end{equation}
\section{Results and discussions}
\subsection{Dataset generation using CFD simulation}
\begin{figure}[b!]
\vspace{-0.5em}
\centering
\includegraphics[width=0.46\textwidth]{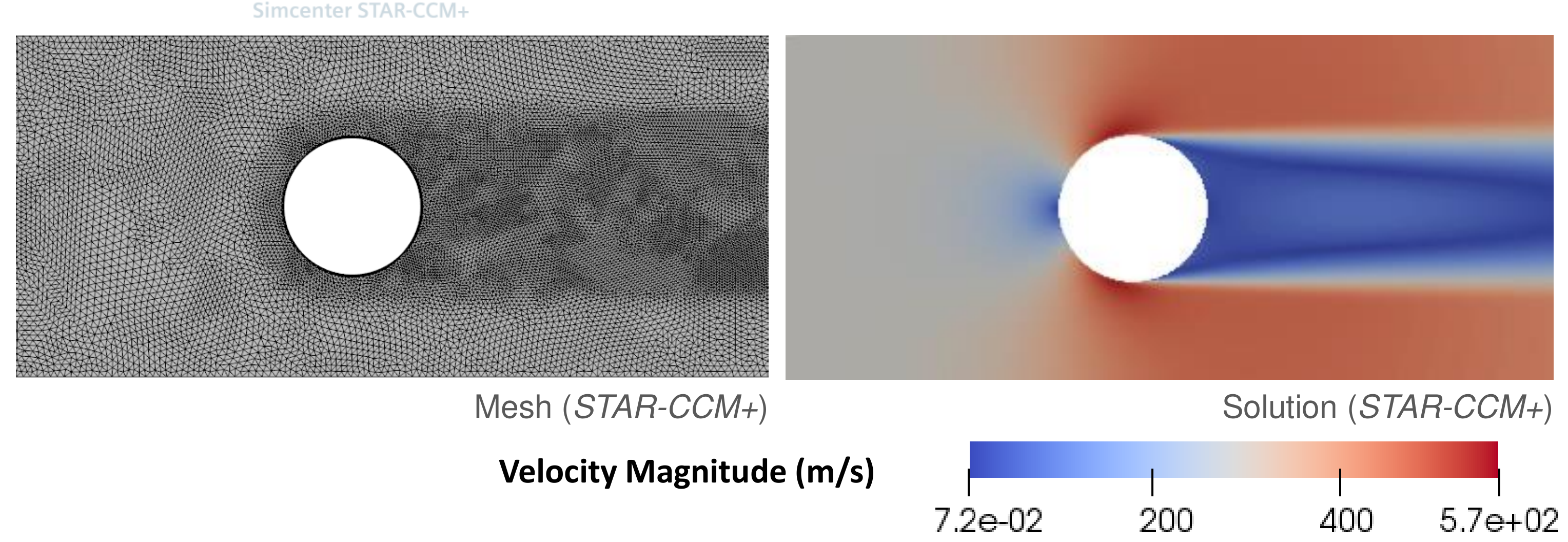}
\caption{\small{Partial view of mesh for flow over a cylinder with refinement at the cylinder surface and wake regions. This mesh is also used for point cloud sampling in PINN training, and takes into account the density variations. Velocity profile shows gradients in regions of refinement.}}
\label{Mesh_sim}
\vspace{-2em}
\end{figure}
\begin{figure}[t!]
\centering
\includegraphics[width=0.43\textwidth]{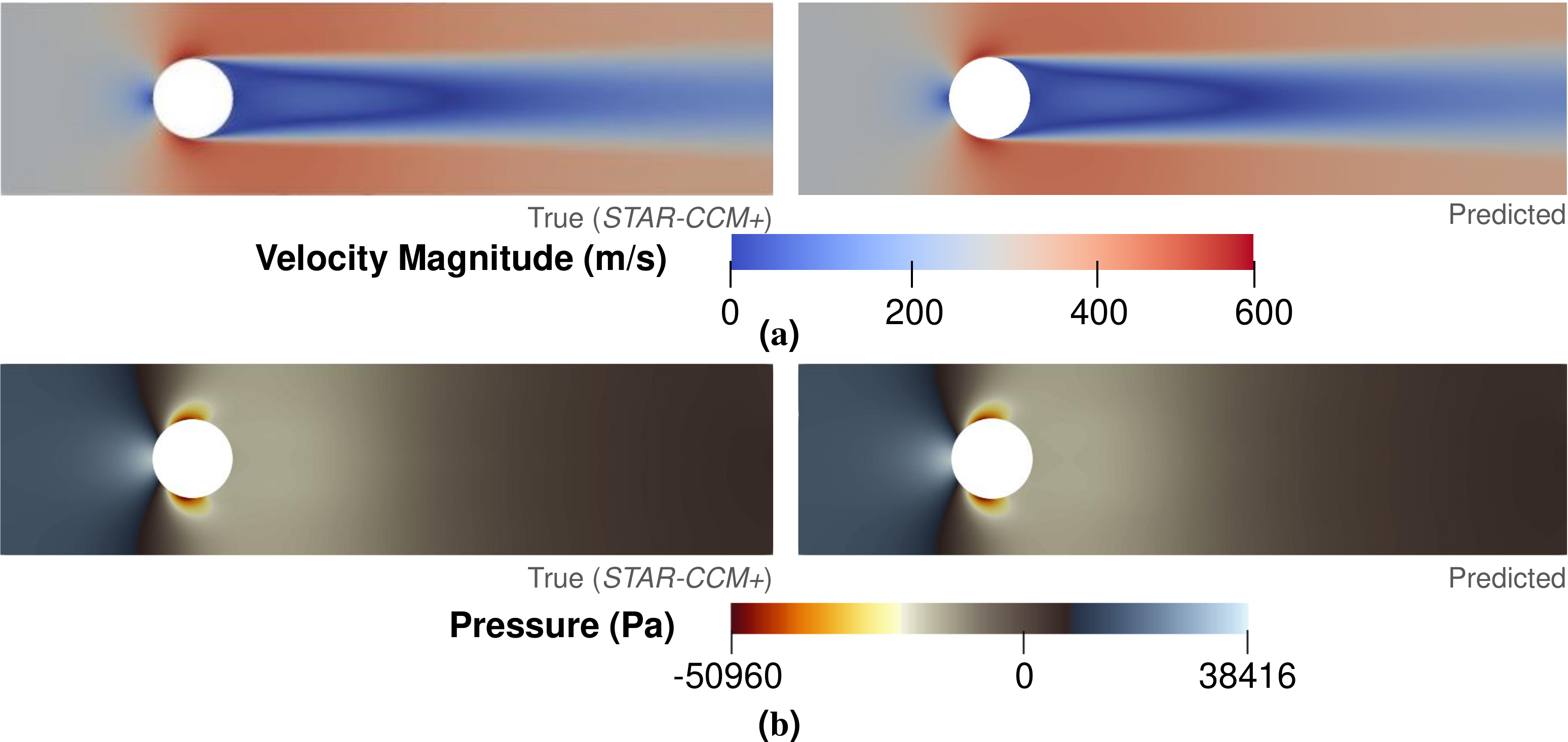}
\caption{\small{Spatial distribution of (a) \textit{velocity magnitude} and (b) \textit{pressure} for a $Re=5600$ flow over the cylinder.}}
\vspace{-1em}
\label{Vel_logloss}
\end{figure}
In this study, we employ Simcenter STAR-CCM+ (\textit{Release 17.02.008}) to simulate turbulent flow scenarios using RANS CFD with the $k$-$\epsilon$ turbulence model. Automatic meshers have been used for each case, with refinement near walls for low wall $y+$, and wall functions for turbulence quantities. Moreover, we have used wake refinements to simulate flow around the cylinder (Fig.~\ref{Mesh_sim}) and the airfoil. The data generated from the simulation is then normalized using the non-dimensional version of the underlying dynamics (i.e., continuity, Navier-Stokes, and RANS equations). We bring the range of various variables to a comparable order of magnitude by normalizing the spatial coordinates, the velocity, and the pressure with the characteristic length, the inlet velocity, and the dynamic pressure, respectively. Later, the data is denormalized again before visualization.
\begin{figure}[b!]
\centering
\includegraphics[width=0.43\textwidth]{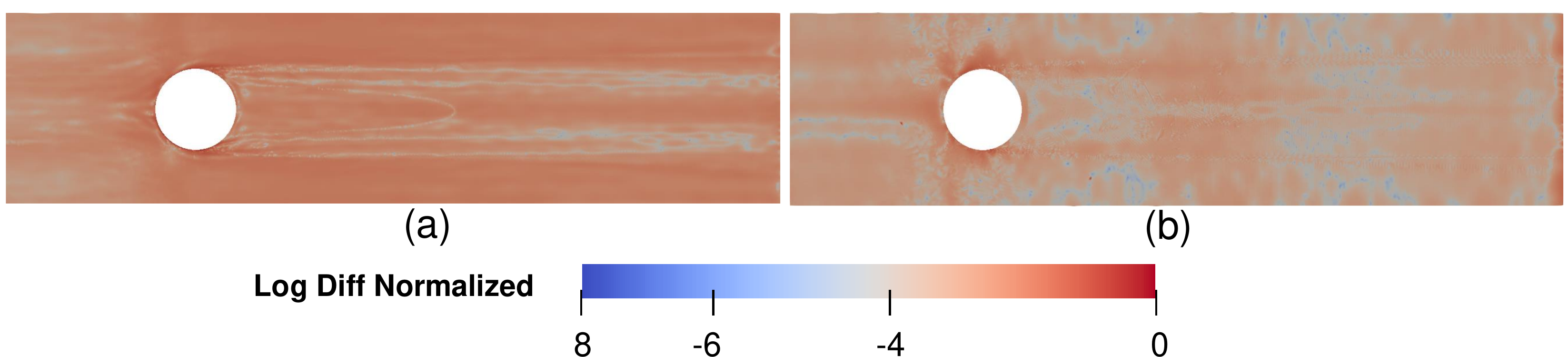}
\caption{\small{Prediction error in (a) \textit{velocity} and (b) \textit{pressure}. To highlight the prediction error, we use normalized values of the logarithm of difference between the true and the predicted values.}}
\label{Pressure_vel_logloss}
\vspace{-1.5em}
\end{figure}
\begin{figure}[t!]
\centering
\label{ablation_fig}
\includegraphics[width=0.43\textwidth]{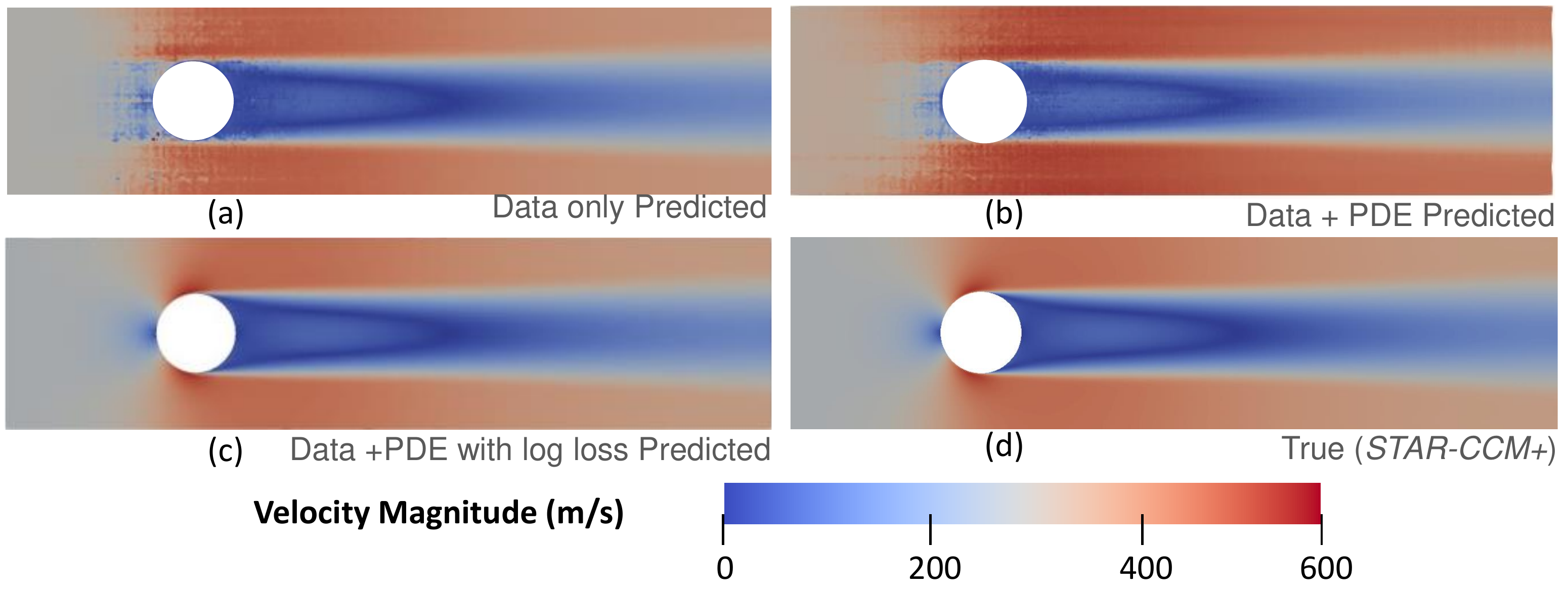}
\caption{\small{Impact of various choices for the loss function. This figure compares the magnitude of the velocity predicted by a PINN trained with (a) only data loss, (b) both data and PDE loss, and (c) both data and PDE loss with a log-loss for $\epsilon$ against its true value from STAR-CCM+ simulation.}}
\label{data_only}
\vspace{-1.5em}
\end{figure}
\subsection{Flow over a cylinder}
While the primary objective of this work is to construct a parametric PINN capable of accommodating varying Reynolds numbers ($Re$), an initial investigation is conducted using single CFD cases (at a fixed $Re$) to assess the optimal training regime. Flow over a cylinder is a well-studied problem in CFD, for both laminar and turbulent flows. The cylindrical obstacle causes a stagnation zone, and the flow diverts around the obstacle. As a result, flow separation occurs and vortex shedding can be seen in the wake. However, steady RANS models average out the periodic unsteady behaviour, resulting in the time averaged flow field. In this work, we employ a constant velocity inlet, along with symmetry planes on the top and bottom walls and a zero pressure outlet. For training, 3000 spatially distributed CFD data points are randomly sampled, with an additional 3000 points dedicated to PDE losses. Fig.~\ref{Vel_logloss}a illustrates the comparison between true and predicted velocity fields, showcasing the aforementioned flow phenomena. The pressure plots (Fig.~\ref{Vel_logloss}b) show a high pressure stagnation region as well as the low pressure flow separation region in both the true and predicted cases.  The differences between true and predicted velocities and pressure for the log-loss training case are shown in Fig.~\ref{Pressure_vel_logloss}. Major losses occur around the cylinder walls, which is known to be a challenging region for all turbulence models due to steep gradients. Moreover, the challenges with using only the data loss or the data+PDE loss but without the logarithmic loss function for $\epsilon$ are highlighted in Fig.~\ref{data_only}. These choices for the loss function yield flow fields with discontinuities and noise stemming from the combination of data and physics losses. This is further reflected in the validation error values reported in Table~\ref{cylinder_vals}. In conclusion, the proposed training regime for RANS-PINN exhibits lower validation losses as well as superior predictive performance.
\begin{table}[h]
\caption{\small{Validation errors for flow over cylinder}}
\vspace{-1em}
\label{cylinder_vals}
\vskip 0.15in
\begin{center}
\begin{small}
\begin{sc}
\begin{tabular}{lcccr}
\toprule
Loss Function & x vel & y vel & Pressure
\\
\midrule
Data Only   & 0.205 & 0.284 & 0.029 \\
Data+PDE   & 0.187 & 0.474 & 0.066 \\
Data+PDE w/ Log-loss & 0.014 & 0.03 & 0.105 \\
\bottomrule
\end{tabular}
\end{sc}
\end{small}
\end{center}
\vspace{-1em}
\end{table}
\subsection{Test on other geometries: Flow over a backwards facing step and NACA 2412 airfoil at a single $Re$}
\begin{figure}[t!]
\centering
\includegraphics[width=0.45\textwidth]{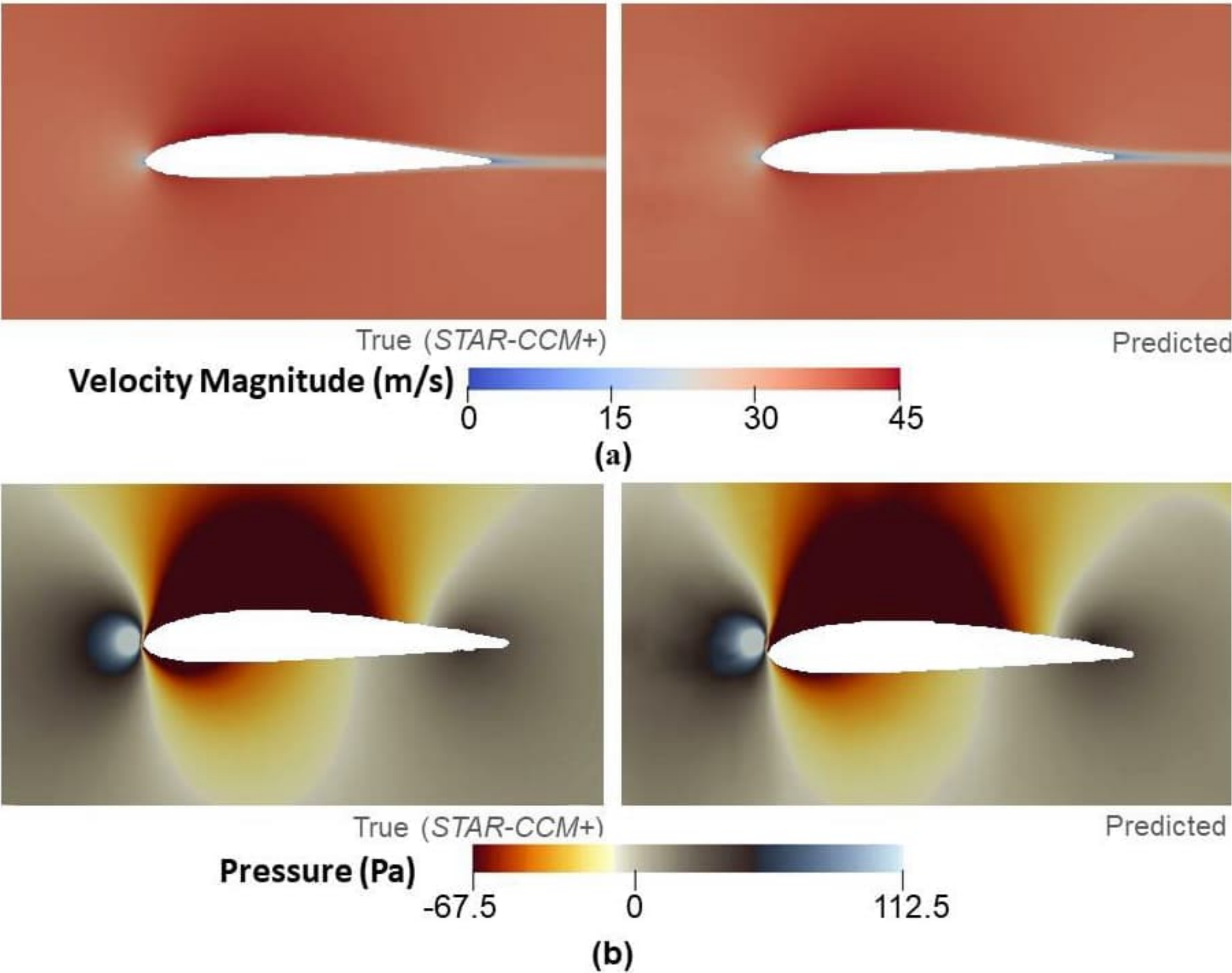}
\caption{\small{Spatial distribution of (a) \textit{velocity magnitude} and (b) \textit{pressure} for a $Re=3\times10^5$ flow around a NACA 2412 airfoil.}}
\label{newGeom_NACA}
\vspace{-1em}
\end{figure}

\begin{figure}[b]
\centering
\includegraphics[width=0.45\textwidth]{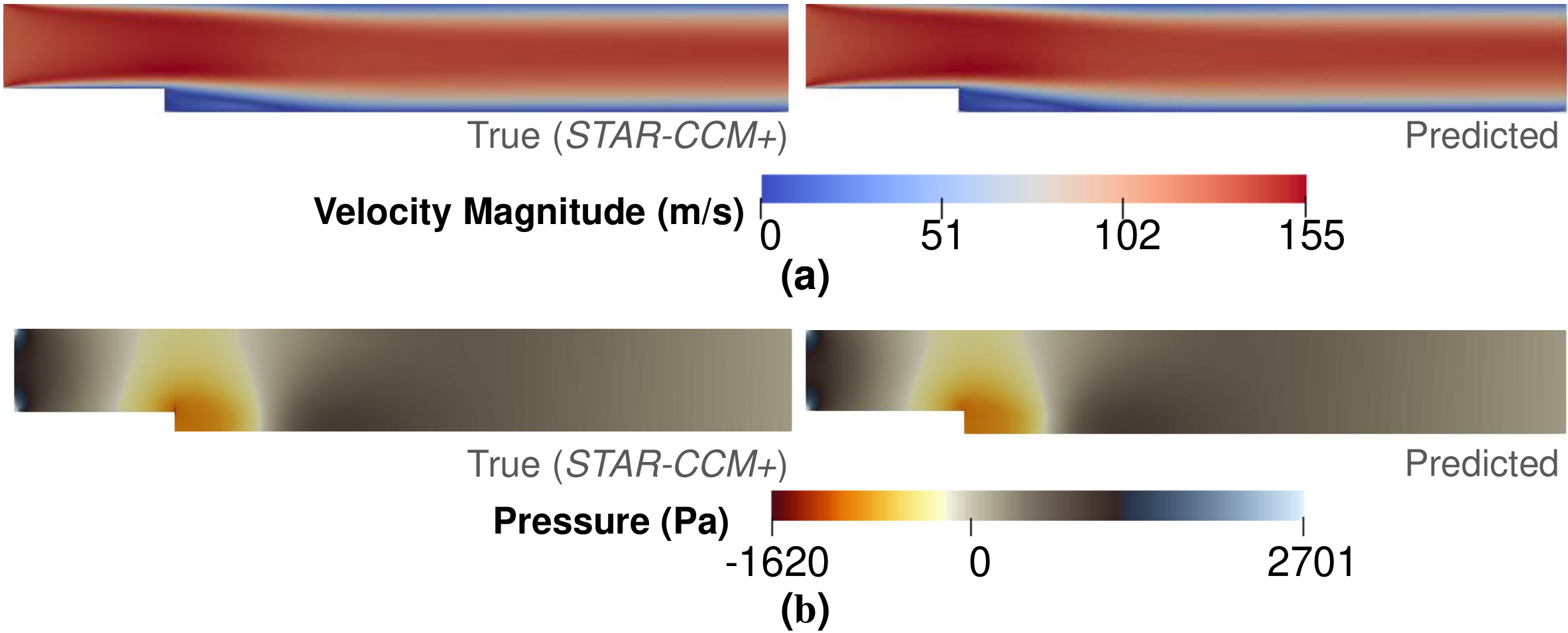}
\caption{\small{Spatial distribution of (a) \textit{velocity magnitude} and (b) \textit{pressure} for a $Re=5600$ flow over a backward facing step.}}
\label{newGeom_BFS}
\vspace{-1.5em}
\end{figure}

\begin{table}[b!]
\vspace{-1em}
\caption{\small{Validation errors for NACA airfoil ($Re=3\times10^5$) and backward facing step ($Re=5600$).}}
\vspace{-1em}
\label{airfoil_BFS_vals}
\vskip 0.15in
\begin{center}
\begin{small}
\begin{sc}
\begin{tabular}{lcccr}
\toprule
Case & x vel & y vel & Pressure \\
\midrule
NACA 2412   & 0.091 & 0.131 & 0.022 \\
Backwards facing step & 0.024 & 0.146 & 0.137 \\
\bottomrule
\end{tabular}
\end{sc}
\end{small}
\end{center}
\vspace{-1em}
\end{table}
To understand the general efficacy of the proposed training method, two additional geometries were chosen for investigation. The first geometry involves airfoils which represents external flows, where a pressure gradient is established between the top and bottom surfaces due to acceleration of flow over the top surface (seen in darker red zones of velocity in Fig.~\ref{newGeom_NACA}a and higher pressure magnitudes in Fig.~\ref{newGeom_NACA}b), which causes lift. The second geometry consists of a backwards facing step (Fig.~\ref{newGeom_BFS}), where a separation bubble form due to sudden expansion in the channel. This leads to flow separation and detachment  and then re-attachment. Both cases had no-slip walls and constant velocity inlet boundary conditions with a zero pressure exit. Low validation error in Table~\ref{airfoil_BFS_vals} and visual inspection of Fig.~\ref{newGeom_NACA} and Fig.~\ref{newGeom_BFS} show that the flow fields have been successfully predicted.

\subsection{Parametric PINN for flow over a cylinder}
After establishing the training regime with these three flow geometries, we revisit the \textit{flow over a cylinder} problem for creating a parametric PINN. The parametric PINN can predict outcomes of CFD simulations for unseen flow scenarios, in particular for any given Reynolds number ($Re$), which depends on the inlet velocity inlet velocities. We achieve this by including the Reynolds number as an additional input to the individual neural networks. 

In this study, we ran CFD simulations for six different Reynolds numbers ranging from 2800 to 5600, with uniform spacing between the values. We sampled 3000 spatial data points from each simulation and utilized them along with PDE losses to train the parametric PINN with $Re$ as the underlying parameter. Although each CFD simulation has 61000 mesh data points, we trained the parametric PINN using only 3000 points, resulting in faster convergence. By leveraging the parametric PINN, we can now predict flow fields for any given Reynolds number. This is highly beneficial for design optimization and exploration studies, as it eliminates the need for additional CFD data to predict primary flow variables across the entire solution domain. Moreover, compared to the traditional approaches, where each CFD simulation run takes approximately 24 core minutes, the parametric PINN can yield results in a near real-time fashion, significantly accelerating the overall process.
\begin{table}[H]
\vspace{-0.5em}
\caption{\small{Generalization error for parametric PINNs (unseen cases)}}
\label{cylinder_vals_parametric}
\vspace{-0.5em}
\begin{center}
\begin{small}
\begin{sc}
\begin{tabular}{lcccr}
\toprule
Case & x vel & y vel & Pressure 
\\
\midrule
$Re=3140$   & 0.139 & 0.289 & 0.164
\\
$Re=5700$   & 0.153 & 0.200 & 0.079 
\\
\bottomrule
\end{tabular}
\end{sc}
\end{small}
\end{center}
\vspace{-2em}
\end{table}

\begin{figure}[t!]
\centering
\includegraphics[width=0.45\textwidth]{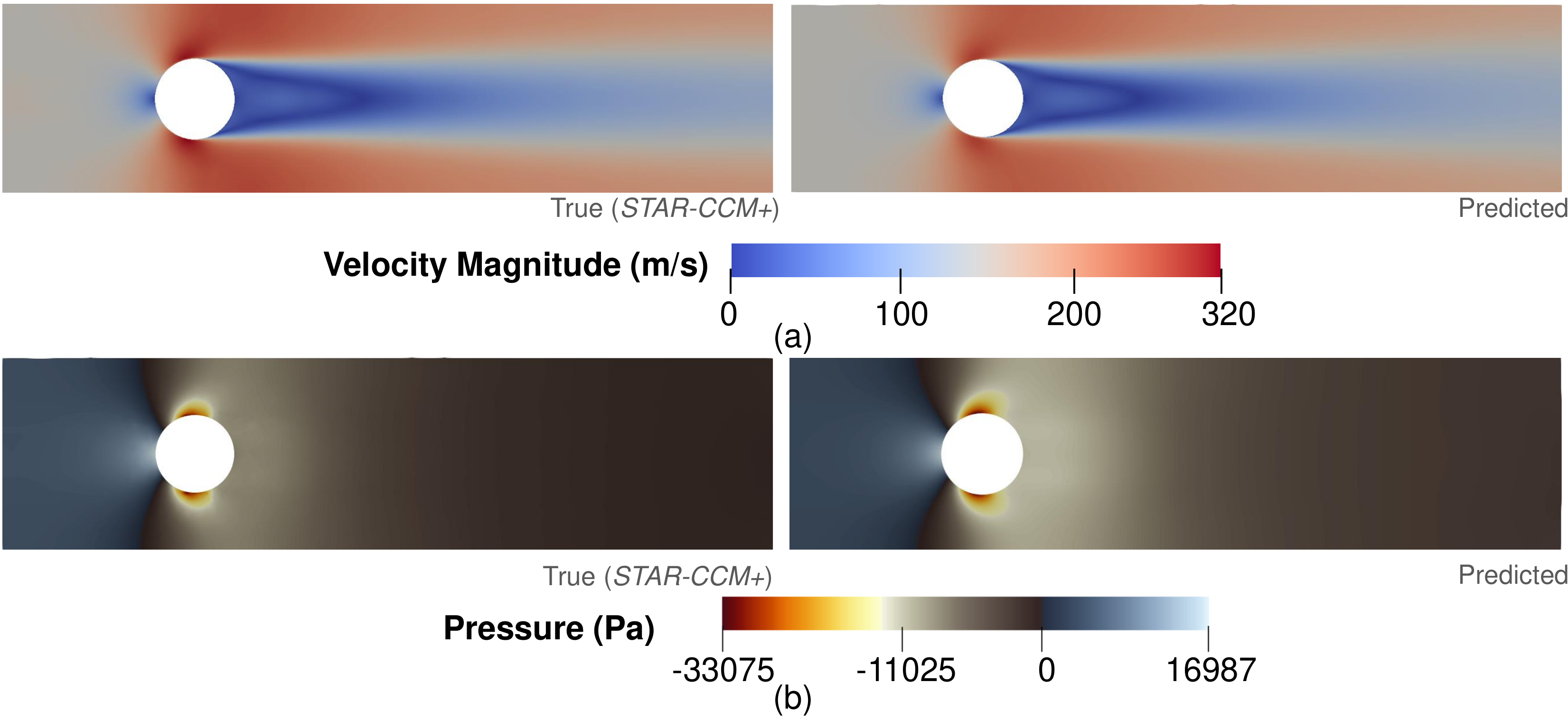}
\caption{\small{Spatial distribution of (a) \textit{velocity magnitude} and (b) \textit{pressure} for a $Re=3140$ flow over the cylinder. We have used a parametric PINN to predict the velocity and the pressure.}}
\label{parametric_PINN_3140}
\vspace{-1em}
\end{figure}

\begin{figure}[b!]
\centering
\includegraphics[width=0.45\textwidth]{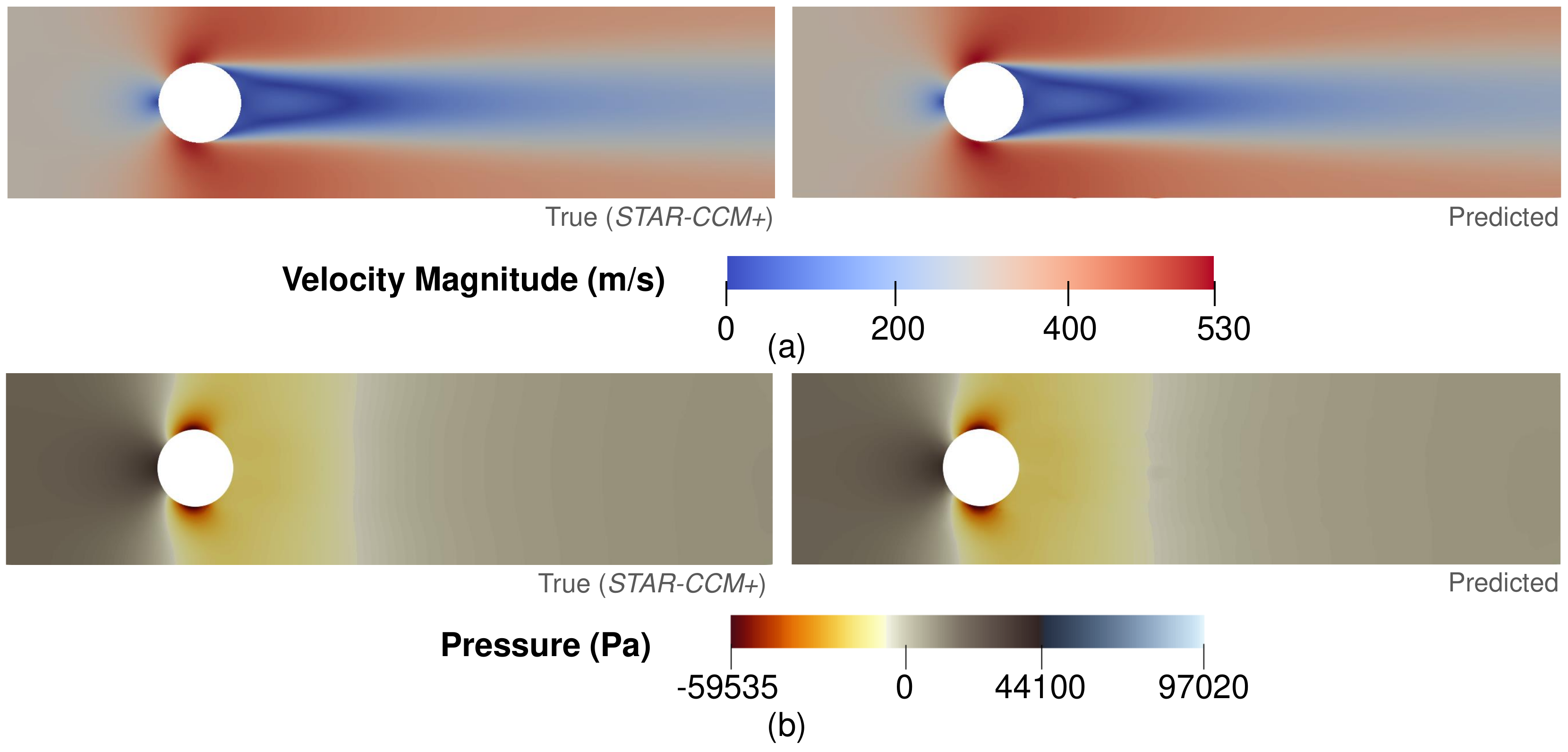}
\caption{\small{Spatial distribution of (a) \textit{velocity magnitude} and (b) \textit{pressure} for a $Re=5700$ flow over the cylinder. We have used a parametric PINN to predict the velocity and the pressure.}}
\label{parametric_PINN_5700}
\vspace{-1em}
\end{figure}

Fig.~\ref{parametric_PINN_3140}a and Fig.~\ref{parametric_PINN_3140}b show the velocity and pressure distributions for $Re=3140$. On the other hand, Fig.~\ref{parametric_PINN_5700}a and Fig.~\ref{parametric_PINN_5700}b display the velocity and pressure distributions for $Re=5700$, which falls outside the training range. In each validation case, we examined 61000 mesh points. Table~\ref{cylinder_vals_parametric} presents the overall error metrics for validation in the case of the parametric PINN.
\section{Conclusion}
PINN-based approaches to learning surrogate models for spatiotemporal systems governed by nonlinear PDEs are relatively common in the literature. However, despite playing an instrumental role in many real-world applications, two-equation RANS turbulence models are yet to be integrated into PINN-based approaches. In this work, we adopt a novel training regime to ensure the successful integration of RANS turbulence model physics into PINNs. Once trained with a limited amount of CFD data, RANS-PINN can yield accurate predictions of overall flow fields for a single Reynolds number. Building upon the successful outcomes of these evaluations for three different flow geometries (flow over a cylinder, a backward-facing step, and a NACA 2412 airfoil), we develop a parametric version of the RANS-PINN to predict flow over a cylinder for any given/unforeseen Reynolds numbers. The parametric RANS-PINN, which highlights how whole simulation cases can be inferred without requiring any CFD data from that specific Reynolds number, offers significant potential in solving design exploration and inverse problems for many real world applications.
\section*{Broader impact}
The current work focuses on turbulent flow problems with two-equation turbulence models. While this perspective is not commonly explored in the PINN literature, these turbulence models hold significant importance in many industrial and academic settings where a lack of computing resources prevents the use of Direct Numerical Simulation (DNS) and Large Eddy Simulation (LES). We can effectively tackle design and inverse problems in many real-world cases by employing a turbulent flow PINN, such as RANS-PINN. The ability to reconstruct a flow field from limited data can help in real-world problems with limited sensor data. Moreover, a parametric PINN trained with minimal CFD data adds significant value to design exploration and optimization by offering a convenient, fast, and computationally efficient means to predict simulation outcomes.
\bibliography{example_paper}
\bibliographystyle{synsml2023}

\newpage
\appendix
\onecolumn
\section{Appendix: Additional Details}

\begin{algorithm}[H]
   \caption{Training Regime}
   \label{alg:example}
\begin{algorithmic}

   \STATE Sample data points for data loss, pde loss and validation from mesh
   \STATE Pretraining individual      NNs using data loss
   \REPEAT
   \STATE Normalize weights of PDE loss components using residual values
   \STATE 
  
   \UNTIL{Loss is stationary}
\end{algorithmic}
\end{algorithm}


The training metrics of the parametric PINN has been showed below in figure \ref{training_pinn}. As seen below, the primary physics errors are less than $10^{-4}$. A brief algorithm of the process has also been shown.
\begin{figure}[H]
\centering
\includegraphics[width=0.8\textwidth]{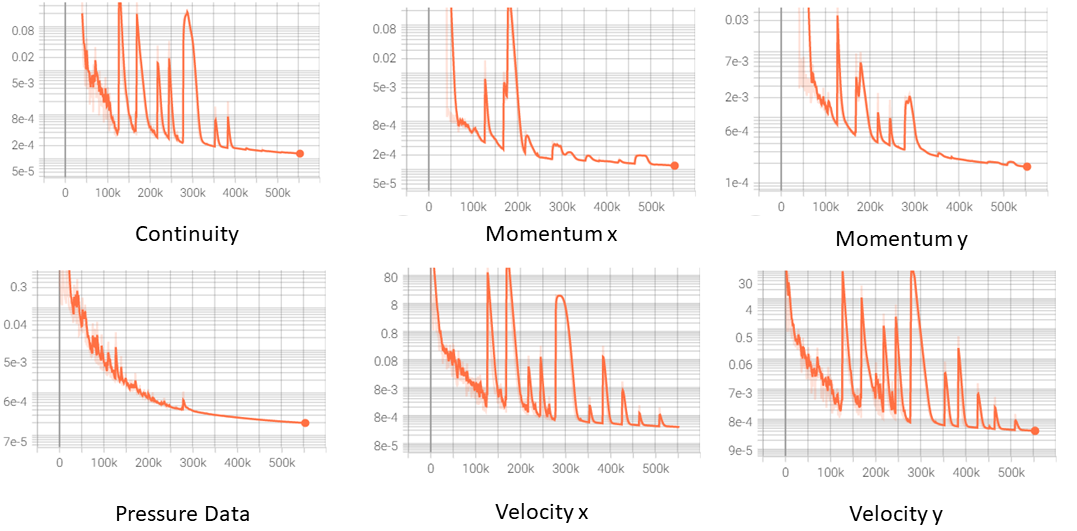}
\caption{Training metrics of parametric PINN}
\label{training_pinn}
\end{figure}


\end{document}